%% file: main.tex
\documentclass[11pt,letterpaper]{article}
\usepackage{emnlp2016}
\usepackage{times}
\usepackage{latexsym}

\usepackage{color}
\usepackage{tabularx}
\usepackage{booktabs}
\usepackage{multirow}
\usepackage{lipsum}
\usepackage{amsmath}
\usepackage{amsfonts}
\usepackage{bigstrut}
\usepackage{arydshln}
\usepackage{url}

\usepackage[]{algorithm2e}

\emnlpfinalcopy

\newcommand{\ignore}[1]{}

\newcommand{\kpgcomment}[1]{\textcolor{blue}{\bf \small [ #1 --KPG]}}
\newcommand{\mbcomment}[1]{\textcolor{red}{\bf \small [ #1 --MB]}}
\newcommand{\damcomment}[1]{\textcolor{green}{\bf \small [ #1 --DAM]}}

\newenvironment{itemizesquish}{\begin{list}{\labelitemi}{\setlength{\itemsep}{-0.2em}\setlength{\labelwidth}{0.5em}\setlength{\leftmargin}{\labelwidth}\addtolength{\leftmargin}{\labelsep}}}{\end{list}}

\title{Who did What: A Large-Scale Person-Centered Cloze Dataset}

\author{Takeshi Onishi\ \ \ \ \ \ Hai Wang\ \ \ \ \ \   Mohit Bansal\ \ \ \ \ \   Kevin Gimpel\ \ \ \ \ \  David McAllester \\
		Toyota Technological Institute at Chicago, Chicago, IL, 60637, USA \\
		\texttt{\{tonishi,haiwang,mbansal,kgimpel,mcallester\}@ttic.edu} \\
                \\
{\small to appear at EMNLP 2016.}}

\date{}

\begin{document}

\maketitle

\begin{abstract}
  We have constructed a new ``Who-did-What'' dataset of over 200,000 fill-in-the-gap (cloze) multiple choice reading comprehension problems
  constructed from the LDC English Gigaword newswire corpus.
  The WDW dataset has a variety of novel features. First, in contrast with the CNN and Daily Mail datasets \cite{anonymized}
  we avoid using article summaries for question formation.  Instead, each problem is formed from two independent articles
  --- an article given as the passage to be read
  and a separate article on the same events used to form the question.  Second, we avoid anonymization --- each choice is a person named entity.
  Third, the problems have been filtered
  to remove a fraction that are easily solved by simple baselines, while remaining 84\% solvable by humans.
  We report performance benchmarks of standard 
  systems and propose the WDW dataset as a challenge task for the community.\footnote{Available at \url{tticnlp.github.io/who_did_what}}
\end{abstract}

\input{./table_sample_questions.tex}

\input{./sec1_introduction.tex}

\input{./sec2_related_work.tex}

\input{./sec3_DataConst.tex}


\input{./sec4_systems.tex}


\input{./sec6_conclusion.tex}


\bibliography{acl2016}
\bibliographystyle{emnlp2016}

\input{./appendix.tex}

\end{document}

%% file: table_sample_questions.tex
\begin{table*}[th]
	\centering
	\fbox{\begin{minipage}[t]{450pt}
{\footnotesize
{\bf Passage:}
Britain's decision on Thursday to drop extradition proceedings against
Gen.~Augusto Pinochet and allow him to return to Chile is
understandably frustrating ... Jack Straw, the home
secretary, said the 84-year-old former dictator's ability to
understand the charges against him and to direct his defense had been
seriously impaired by a series of strokes. ... Chile's
president-elect, Ricardo Lagos, has wisely pledged to let justice run
its course. But the outgoing government of President Eduardo Frei is
pushing a constitutional reform that would allow Pinochet to step down
from the Senate and retain parliamentary immunity from
prosecution. ...

\vspace{1ex}
{\bf Question:} 
Sources close to the presidential palace said that Fujimori declined at the last moment to leave the country and instead he will send a high level delegation to the ceremony, at which Chilean President Eduardo Frei will pass the mandate to XXX.

\vspace{1ex}
{\bf Choices:} (1) Augusto Pinochet (2) Jack Straw (3) Ricardo Lagos

\ignore{
\vspace{2ex}
{\bf Passage:}
U.S. officials said NATO forces would step up efforts to catch
indicted war crimes suspects ... To American statements that NATO
troops would now start patrolling Pale, headquarters of Bosnian Serb
leader and indicted war criminal Radovan Karadzic, the spokesman noted
this had happened for months ...  Tuesday's confusion stemmed from
comments by U.S. Secretary of State Warren Christopher that NATO
troops would increase patrols to give them a better chance of capturing Karadzic
... Nicholas Burns, Christopher's spokesman, said ... He attributed
his information to U.S. Gen. George Joulwan, the NATO commander in
Europe. ...  Simon Haselock, a NATO spokesman in Sarajevo, said
... Karadzic and his military commander, fellow nationalist hardliner
Gen. Ratko Mladic, remain free despite provisions in the Bosnian peace
agreement calling for their arrest. ...

\vspace{1ex}
{\bf Question:} U.S. officials, repeatedly stymied in the quest to oust alleged war criminals from the Bosnian Serb leadership, said Tuesday that increased NATO patrols could improve the chances that indicted suspects such as Radovan Karadzic and XXX will be arrested.

\vspace{1ex} {\bf Choices:} (1) Warren Christopher (2) Nicholas Burns (3) George Joulwan (4) Simon Haselock (5) Ratko Mladic
}

\vspace{2ex}
{\bf Passage:} 
Tottenham won 2-0 at Hapoel Tel Aviv in UEFA Cup action on Thursday
night in a defensive display which impressed Spurs skipper Robbie
Keane. ...  Keane scored the first goal at the Bloomfield Stadium with Dimitar Berbatov, who insisted earlier on Thursday he was happy at the London club, heading a second.
The
26-year-old Berbatov admitted the reports linking him with a move had
affected his performances ...  Spurs manager
Juande Ramos has won the UEFA Cup in the last two seasons ...

\vspace{1ex}
{\bf Question:} Tottenham manager Juande Ramos has hinted he will allow XXX to leave if the Bulgaria striker makes it clear he is unhappy.

\vspace{1ex}
{\bf Choices:} (1) Robbie Keane (2) Dimitar Berbatov
}
\end{minipage}}
\caption{Sample reading comprehension problems from our dataset.}
\label{table:sample}
\end{table*}

%% file: sec1_introduction.tex
\section{Introduction}
Researchers distinguish the problem of general knowledge question answering from that of reading comprehension~\cite{anonymized,Goldilocks}. 
Reading comprehension is more difficult than 
knowledge-based or IR-based question answering in two ways.
First, 
reading comprehension systems must infer answers from a given unstructured passage
rather than structured knowledge sources such as Freebase~\cite{bollacker2008freebase} or the Google Knowledge Graph~\cite{knowedgegraph}.
Second, machine comprehension systems cannot exploit the large level of
redundancy present on the web to find statements that provide a strong syntactic match to the question~\cite{yang2015wikiqa}.
In contrast, a machine comprehension system must use the single phrasing in the given passage, which 
may be 
a poor syntactic match to the question.

\ignore{\mbcomment{After motivating machine comprehension above, we should probably start this next para with first a brief discussion of why current MC datasets are insufficient and hence our new dataset is much needed...}}

In this paper, we describe the construction of a new reading comprehension dataset that we refer to as ``Who-did-What''. 
Two typical examples are shown in Table~\ref{table:sample}.\footnote{The passages here only show certain salient portions of the passage. In the actual dataset, the entire article is given. The correct answers are (3) and (2).}  
The process of forming a problem starts with the selection of a question article from the English Gigaword corpus. 
The question is formed by deleting a person named entity from the first sentence of the question article.
An information retrieval system is then used to select a passage with high overlap with the first sentence of the question article, 
and an answer choice list is generated from the person named entities in the passage.

Our dataset differs from the CNN and Daily Mail comprehension tasks~\cite{anonymized} in that it forms questions from two distinct articles rather than summary points. 
This allows problems to be derived from document collections that do not contain manually-written summaries.
This also reduces the syntactic similarity between the question and the relevant sentences in the passage, increasing the need for deeper semantic analysis.

To make the dataset more challenging we selectively remove problems so as to suppress four simple baselines --- selecting the most mentioned person, the first mentioned person, and two language model baselines.
This is also intended to produce problems requiring deeper semantic analysis.

The resulting dataset yields a larger gap between human and machine
performance than existing ones.  Humans can answer questions in
our dataset with an 84\% success rate compared to the estimates of
75\% for CNN \cite{DBLP:journals/corr/ChenBM16a} and 82\% for the CBT
named entities task \cite{Goldilocks}.  In spite of this higher level of human performance, various existing readers
perform significantly worse on our dataset than they do on the CNN
dataset.  For example, the Attentive Reader \cite{anonymized} achieves 63\% on CNN but only 55\% on Who-did-What
and the Attention Sum Reader 
\cite{AttentionSum} achieves 70\% on CNN but only 59\% on Who-did-What.

In summary, we believe that our Who-did-What dataset is more challenging, and requires deeper
semantic analysis, than existing datasets.



%% file: sec2_related_work.tex
\section {Related Work}
Our Who-did-What dataset is related to several recently developed datasets for machine comprehension.
The MCTest dataset~\cite{richardson-burges-renshaw:2013:EMNLP} consists of 660 fictional stories with 4 multiple choice questions each. 
This dataset is too small to train systems for the
general problem of reading comprehension. 

The 
bAbI synthetic question answering dataset \cite{bAbIICLR16} 
contains passages describing a series of actions in a simulation 
followed by a question. 
For this synthetic data a logical algorithm can be written to solve the problems exactly (and, in fact, is used to generate ground truth answers). 

The Children's Book Test (CBT) dataset, created by \newcite{Goldilocks}, consists of 113,719 cloze-style named entity problems. 
Each problem consists of 20 consecutive sentences from a children's story, a 21st sentence in which a word has been deleted, and a list of ten choices for the deleted word.
The CBT dataset tests story completion rather than reading comprehension. 
The next event in a story is often not determined \---- surprises arise.
This may explain why human performance is 
lower for CBT than for our dataset \---- 82\% for CBT vs.~84\% for Who-did-What.
The 16\% error rate for humans on Who-did-What seems to be largely due to noise in problem formation introduced by errors in
named entity recognition and parsing.  Reducing this noise in future versions of the dataset should significantly improve human performance.
Another difference compared to CBT is that Who-did-What has shorter choice lists on average. Random guessing achieves only 10\% on CBT but 32\% on Who-did-What.
The reduction in the number of choices seems likely to be responsible for the higher performance of an LSTM system on Who-did-What \--- contextual LSTMs
(the attentive reader of Hermann et al., 2015\nocite{anonymized}) improve from 44\%
on CBT (as reported by Hill et al., 2016\nocite{Goldilocks}) to 55\% on Who-did-What.
\ignore{We note that Who-did-What has a larger gap between machine and human performance as measured relative to the gap between random and machine.
\damcomment{This last sentence seems very week --- 10-44-82 is just not very different from 33-58-89.  Machine perfromance is roughly midway between random and human in both cases and the significance of this metric is highly questionable in any case.  I would recommend dropping this.}\kpgcomment{I agree.. I would either move this latter discussion to later in the paper, or drop it.  I think we could include a sentence about how we are primarily interested in measuring the ability to understand paraphrasing relationships in newswire data, while they are mainly measuring the ability to predict what will occur next in a children's story; these tasks are both involved in assessing language understanding and therefore complementary, but they measure distinct capabilities.}}

Above we referenced the 
comprehension datasets 
created
from CNN and Daily Mail articles by \newcite{anonymized}. The CNN and Daily Mail datasets together consist of
1.4 million questions constructed from approximately 300,000 articles.
Of existing datasets, these are the most similar to Who-did-What in that they
consists of cloze-style question answering problems derived from news articles.  As discussed in Section 1, our Who-did-What dataset
differs from these datasets in not being derived from article summaries, in using baseline suppression, 
and in yielding a larger gap between
machine and human performance.  The Who-did-What dataset also differs in that the person named entities are not anonymized, permitting the use of external resources to improve performance while remaining difficult for language models due to suppression.

%% file: sec3_DataConst.tex
\section{Dataset Construction}
\label{sec:details}
We now describe the construction of our Who-did-What dataset in more detail. We sketch the procedure below and provide more specific details in the appendix. 
To generate a problem we first generate the question by selecting a random article --- the ``question article'' --- from the Gigaword corpus
and taking the first sentence of that article --- the ``question sentence'' --- as the source of the cloze question.
The hope is that the first sentence of an article contains prominent people and events which are likely to be discussed in other
independent articles.
To convert the question sentence to a cloze question, we first extract named entities using the
Stanford NER system~\cite{finkel2005incorporating} and parse the sentence using the Stanford PCFG parser~\cite{Klein:2003:AUP:1075096.1075150}.

The person named entities are candidates for deletion to create a cloze problem. 
For each person named entity we then identify a noun phrase in the automatic parse that is headed by that person.  For example,
if the question sentence is ``President Obama met yesterday with Apple Founder Steve Jobs'' we identify the two person noun phrases ``President Obama''
and ``Apple Founder Steve Jobs''.  When a person named entity is selected for deletion,  the entire noun phrase is deleted.  For example, when deleting the second named entity, we get
``President Obama met yesterday with XXX'' rather than ``President Obama met yesterday with Apple founder XXX''. This increases the difficulty of the problems because systems cannot rely on descriptors and other local contextual cues. 
About 700,000 question sentences are generated from Gigaword articles (8\% of the total number of articles).

Once a cloze question has been formed we select an appropriate article as a passage.  The article should be independent of the question article but should discuss the
people and events mentioned in the question sentence.
To find a passage we search the Gigaword dataset using the Apache Lucene information retrieval system~\cite{McCandless:2010:LAS:1893016}, using the question sentence as the query.
The named entity to be deleted is included in the query and required to be included in the returned article.  We also restrict the search to
articles published within two weeks of the date of the question article.
Articles containing sentences too similar to the question in word overlap and phrase matching near the blanked phrase are removed.
We select the best matching article satisfying our constraints.  If no such article can be found, we abort the process and move on to a new question. See the appendix for details.

Given a question and a passage we next form the list of choices.  We collect all person named entities in the passage except unblanked person named entities in the question.
Choices that are subsets of longer choices are eliminated. For example the choice ``Obama'' would be eliminated if the list also contains ``Barack Obama''.
We also discard ambiguous cases where a part of a blanked NE
appears in multiple candidate answers, e.g., if a passage has ``Bill Clinton'' and ``Hillary Clinton'' and the blanked phrase is ``Clinton''. We found this simple coreference rule to work well in practice since news articles usually employ full names for initial mentions of persons.
If the resulting choice list contains fewer than two or more than five choices, the process is aborted and we move on to a new question.\footnote{The maximum of five helps to avoid sports articles containing structured lists of results.}


After forming an initial set of problems we then remove ``duplicated'' problems.  Duplication arises because Gigaword contains many copies of the same article or
articles where one is clearly an edited version of another.  Our duplication-removal process ensures that no two problems have very similar questions. Here, similarity is
defined as
the ratio of the size of the bag of words intersection to the size of the smaller bag.


In order to focus our dataset on the most interesting problems, we remove some problems to suppress the performance of the following simple baselines:
\begin{itemizesquish}
	\item First person in passage: Select the person that appears first in the passage.
	\item Most frequent person: Select the most frequent person in the passage.
	\item $n$-gram: Select the most likely answer to fill the blank under a 5-gram language model trained on Gigaword minus articles which are too similar to one of the questions in word overlap and phrase matching.
	\item Unigram: Select the most frequent last name using the unigram counts from the 5-gram model.
\end{itemizesquish}
To minimize the number of questions removed we 
 solve an optimization problem defined by limiting the performance of each
baseline to a specified target value while removing as few problems as possible, i.e.,
\begin{equation}
	\max_{\alpha(C)} \sum_{C \in \{0,1\}^{|b|}} \alpha(C) | T(C) |
\end{equation}
subject to
\begin{eqnarray}
	&\forall i& \ \sum_{ C : C_i = 1} \frac{\alpha(C) | T(C)| }{N} \le k \nonumber \\
	&N& = \sum_{C \in \{0,1\}^{|b|}} \alpha(C) | T(C) |
\end{eqnarray}
where $T(C)$ is the subset of the questions solved by the subset $C$ of the suppressed baselines, and $\alpha(C)$ is a keeping rate for question set
$T(C)$. $C_i=1$ indicates $i$-th baseline is in the subset and $|b|$ is a number of baselines. Then $N$ is a total number of questions and $k$ is an upper bound for the baselins after suppression. $k$ is set to the random performance.
The performance of these baselines before and after suppression are shown in Table~\ref{table:beforeKnockOff}.
The suppression removed 49.9\% of the questions. 

Table~\ref{table:statistics} shows statistics of our dataset after suppression.  We split the final dataset into train, validation, and test by taking the validation and
test to be a random split of the most recent 20,000 problems as measured by question article date.  In this way there is very little overlap in semantic subject matter between the training set and either validation or test.
We also provide a larger ``relaxed'' training set formed by applying less baseline suppression (a larger value of $k$ in the optimization).  The relaxed training set then has a slightly different distribution from the train, validation, and test sets which are all fully suppressed.

\ignore{
For initial problem set $Q$, we classifies questions $q \in Q$ by its difficulty, e.g. the set of baseline systems which can solve the question, and defined subsets of questions as $T \subset Q$ for each size $m$ of the multiple choice answer set:
\begin{equation}
	T(C) =  \{ q \in Q | \mathbb{I}(\text{ $b_i$ solves $q$}) = C_i \}
\end{equation}
where $C \in \{0,1\}^{|b|}$ and $b$ is the above list of baseline systems. Each subsets corresponds to the difficulty.

We removed questions in each subset $T(C)$ with probabilities $\alpha(C)$ which is a solution of the following optimization problem for each size of $m$:
\begin{equation}
	\max_{\alpha(C), C\in \{0,1\}^{|b|}} \sum_{C \in \{0,1\}^{|b|}} \alpha(C) | T(C) |
\end{equation}

subject to
\begin{eqnarray}
	&\forall i& \ \sum_{ C : C_i = 1} \frac{\alpha(C) | T(C)| }{N} \le k \nonumber \\
	&N& = \sum_{C \in \{0,1\}^{|b|}} \alpha(C) | T(C) |
\end{eqnarray}
where $\sum_{ C : C_i = 1} \frac{\alpha(C) | T(C)|}{N}$ is a new accuracy of a baseline $b_i$ after removing trivial questions and $k \in [0,1]$ is the target accuracy which all new baseline accuracies are intended to be. We employed the random chance to the target accuracy and relaxed it with 12\% for data size, e.g. $k=1/m + 0.12$.
}

\begin{table}[t]
\centering
\begin{tabular}{@{}l|ll@{}}
  \toprule
        & \multicolumn{2}{c}{Accuracy}                          \\
\multicolumn{1}{c|}{Baseline} & \multicolumn{1}{c}{Before} & \multicolumn{1}{c}{After} \\ \hline
First person in passage                &  0.60   &    0.32                      \\
Most frequent person              &  0.61   &    0.33                       \\
$n$-gram                         &  0.53   &    0.33                       \\
Unigram                       &  0.43   &    0.32                      \\
Random$^\ast$ & 0.32 & 0.32 \\
\bottomrule
\end{tabular}
\caption{Performance of suppressed baselines. $^\ast$Random performance is computed as a deterministic function of the number of times each choice set size  appears. Many questions have only two choices and there are about three choices on average. 
}
\label{table:beforeKnockOff}
\end{table}

\begin{table}
\centering
\begin{tabular}{@{}lrrrr@{}}
\toprule
& relaxed & train & valid & test   \\
& train & & & \\ \midrule
\# queries         & 185,978     & 127,786   & 10,000  & 10,000  \\
Avg \# choices     & 3.5   & 3.5      & 3.4    & 3.4    \\
Avg \# tokens      & 378 & 365    & 325  & 326  \\
Vocab size         & 347,406 & \multicolumn{3}{c}{308,602} \\
\bottomrule
\end{tabular}
\caption{Dataset statistics.}
\label{table:statistics}
\end{table}

%% file: sec4_systems.tex
\newcommand{\argmax}{\operatornamewithlimits{argmax}}

\section{Performance Benchmarks}
\label{sec:baselines}
We report the performance of several systems to characterize our dataset:
\begin{itemizesquish}
	\item Word overlap: Select the choice $c$ inserted to the question $q$ which is the most similar to any sentence $s$ in the passage, i.e., ${\rm CosSim}({\rm bag}(c+q), {\rm bag}(s))$.
	\item Sliding window and Distance baselines (and their combination) from \newcite{richardson-burges-renshaw:2013:EMNLP}.
	\item Semantic features: NLP feature based system from \newcite{wang-EtAl:2015:ACL-IJCNLP2}.
	\item Attentive Reader: LSTM with attention mechanism \cite{anonymized}.
	\item Stanford Reader: An attentive reader modified with a bilinear term \cite{DBLP:journals/corr/ChenBM16a}.
	\item Attention Sum (AS) Reader: GRU with a point-attention mechanism \cite{AttentionSum}.
	\item Gated-Attention (GA) Reader: Attention Sum Reader with gated layers \cite{DBLP:journals/corr/DhingraLCS16}.
\end{itemizesquish}

\ignore{
	\item N-gram : Select the most likely candidate answer with n-gram trained on "???" on the training set.
	\item Uni-gram : Select the most likely candidate answer with uni-gram trained on "???" on the training set.
}

\ignore{
{\bf Word overlap: } Select the choice $c$ maximizing the maximum over sentence $s$ in the passage of $\mathrm{overlap}(q(c),s)$
where $q(c)$ is the question with answer $c$ inserted into the blank.

\begin{equation}
	\hat{a} = \argmax_{c \in A} \max_{s \in P} \;\mathrm{overlap}(q(c),s))
\end{equation}
where $q(c)$ is a question whose blank is filled by a candidate choice $c$.
}

\noindent Table~\ref{table:baseline} shows the performance of each system on the test data. For the Attention and Stanford Readers, we anonymized the Who-did-What data by replacing named entities with entity IDs as in the CNN and Daily Mail datasets. 

We see consistent reductions in accuracy when moving from CNN to our dataset. The Attentive and Stanford Reader drop by up to 10\% and the AS and GA reader drop by up to 17\%. The ranking of the systems also changes. 
In contrast to the Attentive/Stanford readers, the AS/GA readers explicitly leverage the frequency of the answer in the passage, a heuristic which appears beneficial for the CNN and Daily Mail tasks. Our suppression of the most-frequent-person baseline appears to more strongly affect the performance of these latter systems. 

\begin{table}
\centering
\begin{tabular}{@{}lrr@{}}
\toprule
System & WDW & CNN\\ \midrule
Word overlap & 0.47 & -- \\
Sliding window & 0.48 & -- \\
Distance & 0.46 & -- \\
Sliding window + Distance  &  0.51 & -- \\
Semantic features & 0.52 & -- \\ \hdashline 
Attentive Reader & 0.53 & $0.63^{I}$ \\
Attentive Reader (relaxed train) & 0.55 &  \\ 
Stanford Reader & 0.64 & $0.73^{I\hspace{-.1em}I}$ \\
Stanford Reader (relaxed train) & 0.65 &  \\
AS Reader & 0.57 & $0.70^{I\hspace{-.1em}I\hspace{-.1em}I}$ \\ 
AS Reader (relaxed train) & 0.59  &  \\
GA Reader & 0.57 & $0.74^{I\hspace{-.1em}V}$ \\
GA Reader (relaxed train) & 0.60 &  \\ \hline 
Human Performance & $84/100$ & $0.75+^{I\hspace{-.1em}I}$ \\ \bottomrule
\end{tabular}
\caption{System performance on test set. Human performance was computed by two annotators on a sample of 100 questions. Result marked ${I}$ is from (Hermann et al., 2015), results marked ${I\hspace{-.1em}I}$ are from (Chen et al., 2016), result marked $I\hspace{-.1em}I\hspace{-.1em}I$ is from (Kadlec et al., 2016), and result marked $I\hspace{-.1em}V$ is from (Dhingra et al., 2016). }
\label{table:baseline}
\end{table}

%% file: sec6_conclusion.tex
\section{Conclusion}
We presented a large-scale person-centered cloze dataset whose
scalability and flexibility is suitable for neural methods. This dataset is different in a variety of ways from existing large-scale cloze datasets
and provides a significant extension to the training and test data for machine comprehension.

\ignore{
Our dataset is answerable by human readers with xx\% and requires reading comprehension of text which is challenging for machines. 

We published our dataset (www....) to help compare each systems of reading comprehension.
}

%% file: appendix.tex
\section{Appendix}
We include pseudocode for generating questions (Alg.~\ref{alg:question}) and multiple choice answer sets (Alg.~\ref{alg:passage}). 

\input{./persuade_code.tex}


%% file: persuade_code.tex
\begin{algorithm}[]
 \label{alg:question}
 \KwData{an article $A$}
 \KwResult{either  $\mathit{null}$, if no question can be formed from $A$, or a cloze question $q$ and the deleted person named entity (NE) $t$.}
  
  $s \Leftarrow$ the first sentence of the article $A$. \\
  \lIf{ \textbf{not} $ 10 \le |s| \le 120 $}{return $\mathit{null}$ }
  
  $E \Leftarrow$ The set of person NEs $e$ in $s$ such that $e$ contains no more than three words. Named entities sharing a word with an earlier named entity are deleted.\\

  \lIf{ $|E|<2$ }{ return $\mathit{null}$ }
  
  $T \Leftarrow $ constituent parse tree of $s$ \\
  \For{ $e \in E$ starting from the end of $s$ }{
  	$b \Leftarrow$ The node in $T$ for person NE $e$. \\
  	\While{ $b.\mathit{category} \in \texttt{ \{NP, NNP, NNPS\}}$ and  $b.\mathit{head} = e$ and no element of $b.\mathit{descendant}$ has category $\texttt{SBAR}$}{
  		$b \Leftarrow b.\mathit{parent}$
  	}
  	\If{no element of $b.descendant$ has head word ``and''}{
          return $(q,e)$ where $q$ is the cloze question formed from deleting $b$ from $s$.
  	}
  }
  return $\mathit{null}$ \\
  ~ \\
  \caption{Question Formation. Named entities are recognized by
    the Stanford NER system and parse trees are generated by the Stanford PCFG
    parser. Here $X.category$ is the syntactic category of parse node $X$, $X.parent$ is the parent-node of the node $X$, $X.descendant$ is
        the set of descendants of $X$ and $X.head$ is the head word of $X$.}
\end{algorithm}

\begin{algorithm}[]
 \label{alg:passage}
{\small 
 \KwData{a pair $(q,e)$ returned by Algorithm 1.}
 \KwResult{either $\mathit{null}$, if no appropriate passage can be found, or a passage $a$ and multiple choice answer set $C$.}
 
 $p \Leftarrow \mathit{null}$
 
 \For{ $a \in \mathit{RankedArticles}(q,e)$}{
 	$C \Leftarrow$ The set of person NEs in $a$ different from $e$ and not in $q$. Named entities appearing as sub-part of an earlier named entity are deleted. \\ 	
   \lIf{ $2 \leq |E| \leq 5$}{return ($a,C$)}
 }
 return $\mathit{null}$
~ \\
~ \\
 
\textit{\textbf{RankedArticles}}$(q,e)$\{ \\
 $A_r \Leftarrow \emptyset$\\
 \For{$t \in \{1,3,7,14\}$}{
 	$A \Leftarrow \mathit{Articles}(q,e,t)$ \\
 	\For{$a \in A$}{
 		\If{ $\mathit{isValid}(a,q)$}{$A_r \Leftarrow A_r$ followed by $a$}
 	}
 }
return $A_r$\} 
\\
~ \\

\textit{\textbf{Articles}}($q,e,t$)\{ \\
\KwResult{articles containing the person NE $e$, published within $t$ days of the article from which $q$ was taken, and ranked by Apache Lucene.\}}
~ \\

\textit{\textbf{isValid}}($a,q$)\{ \\
\ \  $a$ is a valid passage for $q$ if the following hold:
 \begin{itemizesquish}
 	\item no sentence in $a$ shares more than 78\% of its words with the question $q$.
	\item no sentence in $a$ contains the sequence of five words to the left of the blank in $q$, and similarly
          for the sequence to the right.
	\item $a$ contains at least one of the person NEs in $q$. (All person NEs in $q$ are different from $e$. Two
          named entities are considered the same if they share some words.) \}
 \end{itemizesquish}
%
}
~ \\
 \caption{Passage Selection}
\end{algorithm}